\documentclass[letterpaper]{article} 
\usepackage{aaai25}  
\usepackage{times}  
\usepackage{helvet}  
\usepackage{courier}  
\usepackage[hyphens]{url}  
\usepackage{graphicx} 
\usepackage{natbib}  
\usepackage{caption} 
\usepackage{algorithm}
\usepackage{algorithmic}

\usepackage{array}       
\usepackage{amssymb}
\usepackage{pifont}

\usepackage{booktabs}  
\usepackage{multirow}
\usepackage{booktabs}
\usepackage{amsmath}  
\usepackage{xcolor}

\usepackage{newfloat}
\usepackage{listings}
\DeclareCaptionStyle{ruled}{labelfont=normalfont,labelsep=colon,strut=off} 
\floatstyle{ruled}
\newfloat{listing}{tb}{lst}{}
\floatname{listing}{Listing}
\title{InstructOCR: Instruction Boosting Scene Text Spotting}
\author{
    Chen Duan\textsuperscript{\rm 1},
    Qianyi Jiang\textsuperscript{\rm 1},
    Pei Fu\textsuperscript{\rm 1}\thanks{Corresponding author.},
    Jiamin Chen\textsuperscript{\rm 2},
    Shengxi Li\textsuperscript{\rm 1},
    Zining Wang\textsuperscript{\rm 1},
    Shan Guo\textsuperscript{\rm 1},
    Junfeng Luo\textsuperscript{\rm 1}
}
\affiliations{
    \textsuperscript{\rm 1}Meituan\\
    \textsuperscript{\rm 2}Xi'an Jiaotong University\\

    \{duanchen02,jiangqianyi02,fupei,lishengxi,wangzining03,guoshan05,luojunfeng\}@meituan.com\\
    \{chenjm\}@stu.xjtu.edu.cn

}

\usepackage{bibentry}

\begin{document}

\maketitle

\begin{abstract}
In the field of scene text spotting, previous OCR methods primarily relied on image encoders and pre-trained text information, but they often overlooked the advantages of incorporating human language instructions.
To address this gap,  we propose InstructOCR, an innovative instruction-based scene text spotting model that leverages human language instructions to enhance the understanding of text within images. Our framework employs both text and image encoders during training and inference, along with instructions meticulously designed based on text attributes. This approach enables the model to interpret text more accurately and flexibly. 
Extensive experiments demonstrate the effectiveness of our model and we achieve state-of-the-art results on widely used benchmarks. 
Furthermore, the proposed framework can be seamlessly applied to scene text VQA tasks. By leveraging instruction strategies during pre-training, the performance on downstream VQA tasks can be significantly improved, with a 2.6\% increase on the TextVQA dataset and a 2.1\% increase on the ST-VQA dataset. These experimental results provide insights into the benefits of incorporating human language instructions for OCR-related tasks. 

\end{abstract}

 \begin{links}
     \link{Code}{https://github.com/ChenD-VL/InstructOCR}
\end{links}

\begin{figure}[h]
\centering
  \includegraphics[width=0.39\textwidth]{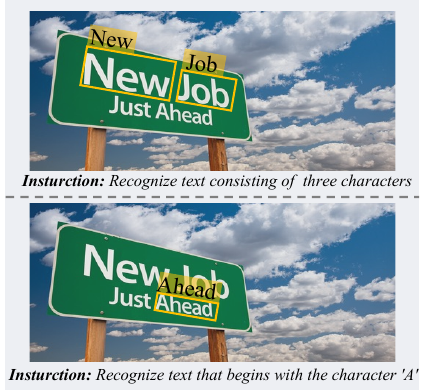}
  \caption{Examples of text recognition results generated using various instructions. This illustrates how different instructions can influence the output of the text recognition process.}
  \label{fig:teaser}
\end{figure}

\section{Introduction}
Scene text spotting technology aims to detect and recognize characters directly within natural scene images. Recently, significant advancements in integrating vision and text have been made across various visual-language tasks, leading to the emergence of innovative instruction-based models. Some studies \cite{groundingdino, geng2023instructdiffusion,brooks2023instructpix2pix,sam} have validated that incorporating human language instructions can enable models to comprehend the content of images more accurately. 
Drawing on the insights gained from these studies, we pose the following question: \textbf{For scene text images, which inherently involve visual text, wouldn't the incorporation of human language instructions be more beneficial?}


 
In OCR tasks, several works, including oCLIP \cite{oclip2022} and ODM \cite{odm}, have validated the effectiveness of pretraining the image encoder with relevant text information. However, these methods only employ the image encoder for text spotting tasks without utilizing the text encoder. TCM \cite{yu2023turning} and FastTCM \cite{yu2024turning} extract image and text-based prior knowledge through visual prompt learning and cross-attention within CLIP \cite{radford2021learning}. Nonetheless, the text encoder used in their works is from CLIP and is frozen during their training process. STEP\cite{garcia2024step} attempts to control recognition through regular expressions. However, this method can only process numbers or letters.

We believe that aligning human language instructions with visual text will undoubtedly be advantageous for scene text spotting tasks.
To substantiate this point, we propose a novel framework named InstructOCR, an instruction-based scene text spotting model that facilitates interaction through human language. Our objective is to leverage both text and image encoders during training and inference via instructions. To this end, we design a text encoder that extracts linguistic features from instructions to enhance the model's understanding of text within images. Figure \ref{fig:teaser} illustrates the flexibility of InstructOCR in producing distinct detection and recognition results corresponding to varying input instructions.

\begin{table}[t]
  \centering
  \begin{tabular}{m{0.44\textwidth}}
    \includegraphics[width=\linewidth]{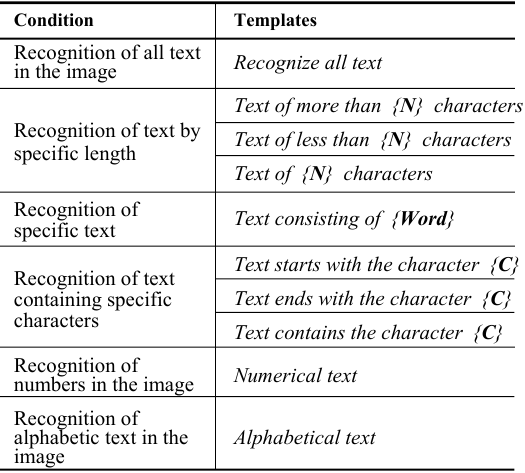}
  \end{tabular}
  \caption{Different templates used in InstructOCR for scene text spotting allow $N$
to be any integer, $C$ to be any letter, and $Word$ to be either a single word or multiple words in the image, resulting in highly diverse instructions.}
  \label{table-image}
\end{table}

The foundation of InstructOCR is built upon a series of instructions derived from annotated information in existing datasets, which are related to text attributes. This is because text attributes, such as text length, are crucial for enhancing the performance of models in scene text recognition tasks \cite{xie2019aggregation,du2023context}. Building on this insight, we have designed ten templates based on these text attributes. As shown in Table \ref{table-image}, these templates can be used to randomly generate a wide variety of training instructions. This diversity is vital as it provides clear semantic directives, enabling the model to perform tasks aligned with human intent. Moreover, this approach fully leverages existing annotated information, thereby eliminating the need for additional annotation costs.



The main contributions of this work can be summarized as follows:
\begin{itemize}
  \item We propose InstructOCR, an end-to-end instruction-based scene text spotting model that significantly enhances the model's understanding of visual text through the use of instructions. To the best of our knowledge, this is the first work that innovatively integrates human language instructions into the domain of scene text spotting. Additionally, with the introduction of instructions, the framework can also be seamlessly extended to VQA tasks.
  
  \item We have meticulously designed a range of instructions specifically tailored for scene text domain, which can leverage existing publicly available scene text datasets at no additional cost. By providing precise and diverse textual annotations, our instructions enhance the model's ability to accurately and efficiently interpret the complex variations of text in images, thereby further improving performance metrics in both scene text spotting and VQA tasks.

  \item We conduct extensive experiments to evaluate the effectiveness of our proposed method, and the results show that InstructOCR provides excellent performance on a range of scene text spotting datasets, achieves state-of-the-art (SOTA).
\end{itemize}


\begin{figure*}[t]
\centering
\includegraphics[width=0.88\textwidth]{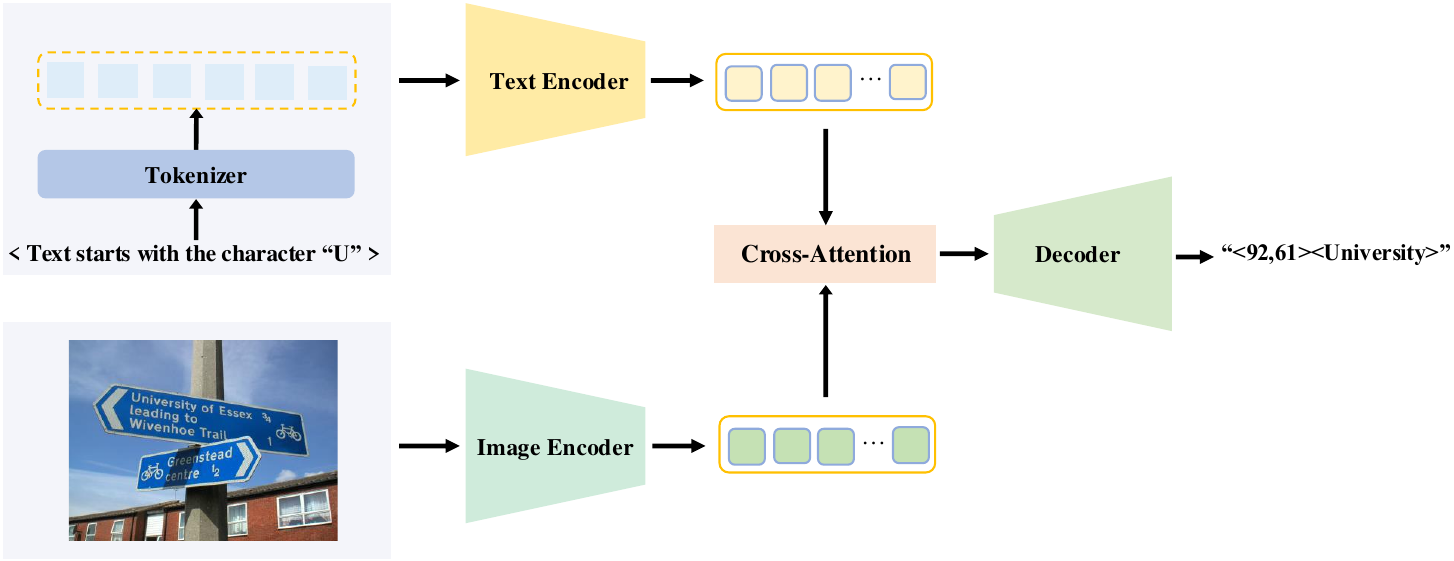}
\caption{Main framework of InstructOCR. InstructOCR is an encoder-decoder architecture, with input branches consisting of an image encoder and a text encoder that handle visual and textual features separately.}
\label{fig2}
\end{figure*}

\section{Related Work}

\subsection{None Sequence-based Method}


Most previous scene text spotting methods have treated detection and recognition as two separate tasks~\cite{borisyuk2018rosetta,liao2017textboxes,huang2022swintextspotter,2021abcnetv2}.
MANGO \cite{qiao2021mango} proposes a one-stage text spotting framework with Mask Attention Guidance, allowing for the direct recognition without the need for RoI operations. The Mask TextSpotter series \cite{2018masktextspotv1,2021masktextspotv2,2020masktextspotv3} leverages the advantages of character-level annotations to achieve character segmentation and can recognize scene text of arbitrary shapes. PGNet \cite{wang2021pgnet} introduces a graph refinement module to optimize coarse recognition and enhance end-to-end performance. Additionally, there is also a parallel mode of detection and recognition. TTS \cite{kittenplon2022towards} proposes a weakly supervised learning method and employs a shared query mechanism for its detector and recognizer. Estextspotter \cite{huang2023estextspotter} and SRSTS \cite{wu2022decoupling} process text detection and recognition in parallel, thereby decoupling text recognition from dependency on detection. Inspired by the DETR \cite{carion2020end} family models. TESTR \cite{zhang2022text} proposes a framework free from Region-of-Interest operations and heuristics-driven post-processing procedures. DeepSolo \cite{ye2023deepsolo} is a DETR-like model that employs a query form with explicit points sampled from the Bezier center curve of text instance lines to efficiently encode the text's position, shape, and semantics.

Similar to scene text spotting methods, early VQA approaches also followed a two-stage task solutions~\cite{singh2019towards,hu2020iterative,yang2021tap}, OCR results and pre-computed features are fed into a vision-and-language model. These approaches lack an interaction between text being recognized and the representation of its context.


\subsection{Sequence-based Method}
Inspired by the immense success in natural language processing, computer vision tasks are converging to Transformers. These approaches extract the image feature along with the transformer decoder to handle various tasks by predicting sequence. Pix2seq V1 and V2~\cite{chen2021pix2seq, chen2022unified} show that it is possible to output boxes and labels as a sequence of discrete tokens, allowing them to have a similar training and decoding interface as language models. ~\cite{kim2021donut} pioneered in employing a prompt-based sequence generation framework for document understanding. 

In the end-to-end scene text spotting methods, the SPTS series ~\cite{peng2022spts,liu2023spts} use the central point of text regions to represent the location and auto-regressively predict coordinate tokens and word transcription tokens. Instead of using separate losses for detection and recognition, ~\cite{kil2023towards} unifies various detection formats, including single point, quadrilaterals, and polygons to a sequence generation paradigm. OmniParser \cite{wan2024omniparser} sets itself apart from other text spotting methods by employing a unified architecture capable of simultaneously addressing text spotting, key information extraction, and table recognition using a single model.
These methods either only utilize image features or do not provide an explicit learnable text encoder, resulting in a lack of understanding of human language instructions and failing to fully leverage human language instructions to enhance the model's understanding of text within images.

Similar to the prevailing trend in scene text spotting tasks, current VQA tasks also primarily utilize sequence-based methods. Particularly, extremely large OCR-free image-text models have shown promising results on VQA tasks (\textit{e.g.,} InternVL~\cite{chen2024internvl}, Monkey~\cite{li2024monkey} and Flamingo~\cite{alayrac2022flamingo}).



\section{Method}
We introduce InstructOCR, an end-to-end scene text spotting method that leverages instructions to guide the model's output and enhance its understanding of the text. This enables the model to generate recognition results based on the provided instructions. The complete process is illustrated in Figure~\ref{fig2}, the instructions are input into the model alongside the image. In the following sections, we will detail the structure of our model.

\subsection{InstructOCR Architecture}

InstructOCR is an encoder-decoder architecture that encompasses a text encoder, an image encoder, and a decoder, designed to synergistically process and interpret textual information within visual contexts.


\textbf{Image Encoder}. The image encoder utilizes the ResNet50 architecture \cite{he2016deep} to extract features from the input image.

\textbf{Text Encoder}. The text encoder adopts BERT \cite{devlin2018bert}, a 12-layer transformer specifically designed to extract features from instructions. By applying cross-attention between the extracted visual and textual features, we obtain instruction-based encoded features. These features are further processed by the decoder to generate the output sequence.

\textbf{Decoder}. The decoder of InstructOCR following the approach of SPTS \cite{peng2022spts}, utilizes an auto-regressive Transformer to generate long sequences for all text instances. Each text instance is represented by a sequence composed of three parts: $[x,y,t]$, where $(x,y)$ represents the center point coordinates and $t$ represents the transcription text. To simplify the representation, the coordinates are uniformly discretized into integers ranging from 1 to 1000. The text is either padded or truncated to a fixed length of 25, and the maximum number of text instances in an image is set to 60. In the sequence representation, the $<PAD>$ token is used to fill in the gaps for shorter text instances. Additionally, $<SOS>$ and $<EOS>$ tokens are inserted at the beginning and end of the sequence, respectively, to indicate the start and end of the sequence. 

\begin{figure}[t]
  \centering
  \includegraphics[width=0.46\textwidth]{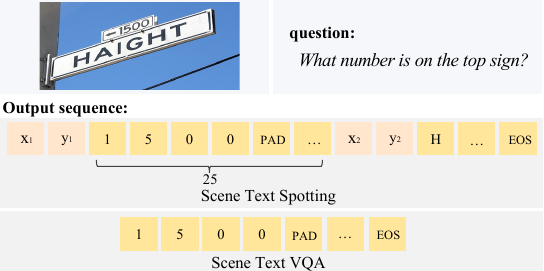}
  \caption{Examples of decoder output for scene text spotting and VQA tasks.}
  \label{fig3output}
\end{figure}

For the VQA task, the question is encoded by the text encoder, while the answer is treated as a direct sequence, as illustrated in Figure~\ref{fig3output}. The entire sequence length is set to 256.

\subsection{Instructions Generation}


Existing OCR annotations typically consist of locations and recognition results. To eliminate the need for human-annotated instructions, we have developed an automated data processing method for generating instructions. Each word can be divided into different attributes, such as whether the initial character is capitalized, whether it contains a certain character, etc. We have designed ten types of templates, as detailed in Table \ref{table-image}, each of which can generate the final instruction through random conditions.

It should be noted that, unlike some methods that use indicators \cite{kil2023towards,peng2023upocr} or specific task prompts \cite{lv2023kosmos,kim2023scob,ye2023ureader} to define different downstream tasks, we use human language as instructions. There are three reasons for this: (1) Our instructions incorporate random numbers and letters, making it impossible to enumerate the types of tasks. (2) Using human language allows us to better leverage the BERT weights pre-trained on large-scale corpora, thereby enabling the model to better understand the semantics of instructions. (3) Pre-training scene text spotting with human language instructions makes it easier to extend its use to other downstream tasks, such as Visual Question Answering (VQA).


During the training process, instructions are selected randomly. The training set is denoted as $\{x_i\}$, where each training datum $x_i$ can be represented in the form of $\{c_i, s_i, t_i\}$. Here, $c_i$ denotes the control instruction, $s_i$ represents the source image, and $t_i$ represents the target text sequence. For a given image, we have a set of annotated text instances denoted as $t' = \{t'_0, t'_1, \ldots, t'_{n-1}\}$. Based on different instructions, we filter out qualifying annotated text instances, resulting in a set denoted as $t = \{t_0, t_1, \ldots, t_{m-1}\}$. The objective of our method is to generate the target text sequence $t_i$ given an input source image $s_i$ and the corresponding instruction $c_i$. For the prediction phase, the instruction is set to ``\textless Recognize all text\textgreater".

\subsection{Loss Function}
In InstructOCR, the training objective is to predict tokens, and we utilize the standard cross-entropy loss for model training. This loss function aims to maximize the likelihood of the correct tokens during training. The mathematical expression of the cross-entropy loss is as follows:
\begin{equation}
\mathcal{L}_{seq} = \text{maximize} \sum_{i=1}^{L} w_i \log P(\tilde{s}_i | I, s_{1:i})
\end{equation}
where $I$ is the input image, $s$ is the input sequence, $\tilde{s}$ is the output sequence, $L$ is the length of the sequence, and $w_i$ is the weight of the likelihood of the $i-th$ token, which is empirically set to 1.

\begin{table*}[t]
  \setlength\tabcolsep{9.8pt}
  \begin{tabular}{ccccccccc}
    \toprule
    \multirow{2}{*}{Methods} & \multicolumn{2}{c}{Total-Text}  & \multicolumn{3}{c}{ICDAR2015} & \multicolumn{3}{c}{ICDAR2013} \\
    \cmidrule(lr){2-3} \cmidrule(lr){4-6} \cmidrule(lr){7-9}
     & None & Full  & S & W & G & S & W & G \\
    \midrule
    \multicolumn{9}{c}{Bounding Box-based methods} \\
    \midrule 
    Mask TextSpotter \cite{2018masktextspotv1} & 65.3 & 77.4 & 83.0& 77.7 &73.5 & 93.3 & 91.3 & 88.2 \\
    FOTS\cite{liu2018fots} & - & -  & 83.6 & 79.1 & 65.3  & 92.0 & 90.1 & 84.8 \\
    Boundary TextSpotter \cite{wang2020all}  & 65.0 & 76.1 &79.7 &75.2 & 64.1 &88.2 &87.7 &84.1 \\
    PGNet \cite{wang2021pgnet} & 63.1 & -  & 83.3 & 78.3 & 63.5 & - & - & - \\
    MANGO \cite{qiao2021mango}& 72.9 & 83.6  & 81.8 & 78.9 & 67.3 & 92.9 &92.7 &88.3 \\
    ABCNet v2 \cite{2021abcnetv2} & 70.4 & 78.1  & 82.7 & 78.5 & 73.0 & - & - & - \\
    SwinTextSpotter\cite{huang2022swintextspotter} & 74.3 & 84.1 & 83.9 & 77.3 & 70.5 & - & - & - \\
    SRSTS\cite{wu2022decoupling} & 78.8 & 86.3  &85.6 & 81.7 & 74.5 & - & - & - \\
    GLASS\cite{ronen2022glass} & 79.9 & 86.2  & 84.7 & 80.1 & 76.3 & - & - & - \\
    TESTR\cite{zhang2022text} & 73.3 & 83.9 & 85.2 &79.4 &73.6 & - & - & - \\
    TTS\cite{kittenplon2022towards} & 78.2 &86.3  & 85.2 &81.7 &77.4 & - & - & - \\
    DeepSolo\cite{ye2023deepsolo} & 82.5 & 88.7 & 88.0 &83.5 &79.1 & - & - & - \\
    ESTextSpotter\cite{huang2023estextspotter} & 80.8 & 87.1  & 87.5 &83.0 &78.1 & - & - & - \\
    UNITS$_{\text{-Swin}}$\cite{kil2023towards} & 78.7 & 86.0& 89.0 & 84.1 & 80.3 & - & - & - \\
    DNTextSpotter\cite{qiao2024dntextspotter} &84.5 &89.8 & 88.7 &84.3&79.9& - & - & - \\
    \midrule
    \multicolumn{9}{c}{Point-based methods} \\
    \midrule
    TOSS\cite{tang2022you} & 65.1 & 74.8  & 65.9 & 59.6 & 52.4 & - & - & - \\
    SPTS\cite{peng2022spts} & 74.2 & 82.4  & 77.5 & 70.2 & 65.8 & 93.3 & 91.7 & 88.5 \\
    SPTS-v2\cite{liu2023spts} & 75.5 & 84.0 & 82.3 & 77.7 & 72.6 & 93.9	 & 91.8 & 88.6 \\
    InstructOCR & 77.1 & 84.1  & 82.5 & 77.1 & 72.1 & 93.3 & 92.4 & 88.8 \\
    InstructOCR$\dagger$ & 83.4 &  88.3  & 87.5 & 84.2 & 80.6 & 94.9 & 94.1 & 91.7 \\
    \bottomrule
  \end{tabular}
\caption{Scene text spotting results on Total-Text, ICDAR2015, and ICDAR2013. ‘None’ means lexicon-free. ‘Full’ indicates that we use all the words that appeared in the test set. ‘S’, ‘W’, and ‘G’ represent recognition with ‘Strong’, ‘Weak’, and ‘Generic’ lexicons, respectively. $\dagger$ denotes the incorporation of TextOCR and HierText datasets in the training data.}
\label{tab:results}
\end{table*}

\section{Experiments}

\subsection{Datasets}
\textbf{Scene Text Spotting}. In our experiments, we evaluate our method on Total-Text~\cite{ch2017total}, ICDAR2015~\cite{karatzas2015icdar}, and ICDAR2013~\cite{karatzas2013icdar}. Total-Text is an arbitrarily shaped word-level scene text benchmark, with 1,255 training images and 300 testing images. ICDAR2015 contains 1,000 training images and 500 testing images for quadrilateral scene text. ICDAR2013 contains 229 training images and 233 testing images with horizontal text.

\textbf{VQA for Scene Text}. Scene text VQA involves answering questions about the natural scene images or reasoning about the scene text. TextVQA~\cite{singh2019towards} contains 45,336 questions on 28,408 images that require reasoning about text to answer. ST-VQA comprises 23, 038 images sourced from a combination of public datasets~\cite{biten2019scene}


\subsection{Implementation Details}
The Transformer encoder and decoder consist of $6$ layers with $8$ heads. The max length of recognition queries is $25$ and the maximum number of objects is $60$. The entire model is distributively trained on $32$ NVIDIA A100-80G GPUs. We pretrain the model on a combination dataset that includes ICDAR2013, ICDAR2015, Total-Text, Curved Synthetic Dataset 150k~\cite{2020abcnet}, and ICDAR2017 MLT~\cite{nayef2017icdar2017}. And the input for the text encoder is a fixed instruction: ``\textless Recognize all text\textgreater". 

We use a batch size of $320$, and the pretrain model is trained for $200$ epochs, with an initial 5-epoch warm-up phase. We use AdamW optimizer with a learning rate of $4.3 \times 10^{-4}$. The input image's short size is randomly resized to a range from $704$ to $1024$ (intervals of $32$), the maximum length of image is set as $1024$. Subsequently, the model is trained for another $40$ epochs, with a fixed learning rate of $1 \times 10^{-4}$, and the maximum length of image is set as $1600$. Then, instructions are added, and the model is further trained for another $50$ epochs. For the scene text spotting task, the model is fine-tuned on the corresponding real datasets for another $140$ epochs, with a fixed learning rate of $1 \times 10^{-5}$. For the scene text VQA task, the model is fine-tuned on the TextVQA and ST-VQA datasets for another 120 epochs.
At the inference stage, we resize the image's maximum length shorter than $1920$ pixels.

\begin{figure*}[t]
  \centering
  \includegraphics[width=0.89\textwidth]{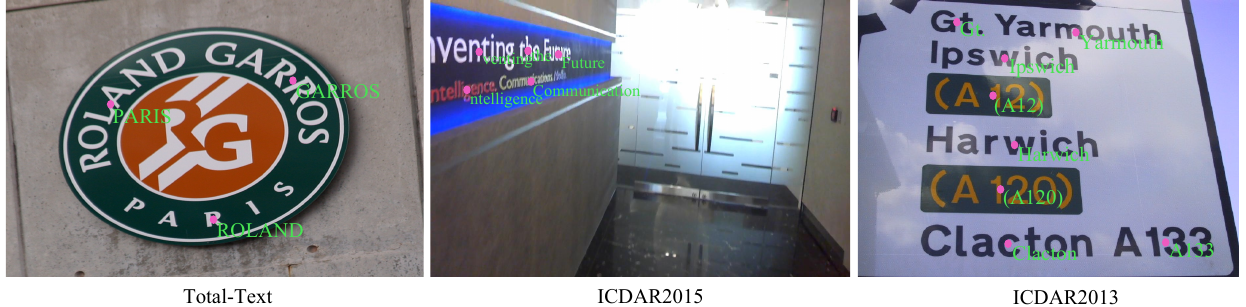}
  \caption{Visual results on Total-Text, ICDAR2015 and ICDAR2013. Our model can effectively handle curved, distorted, and blurred scene text.}
  \label{fig4}
\end{figure*}

\subsection{Comparison with Scene Text Spotting Methods}

We evaluate the model using the point-based metric proposed in SPTS~\cite{peng2022spts}. Notably, our model adeptly outputs the coordinates of single-point and has been compared with other point-based methods. We have also listed methods based on bounding boxes for comparison. However, some of these methods use additional datasets, such as UNITS~\cite{kil2023towards}, which additionally utilizes the TextOCR~\cite{singh2021textocr} and HierText~\cite{long2022towards} datasets. To ensure a fair performance comparison, we have augmented the pre-training with these two additional datasets. Subsequent ablation studies also include these two datasets. Figure \ref{fig4}  displays some exemplary visualization results for scene text spotting.

\textbf{Total-Text: Arbitrarily-Shaped Text}. To validate the generalization ability of our method for arbitrarily shaped scene text spotting, we tested our approach on Total-Text. The scene text spotting results are shown in Table \ref{tab:results}, where InstructOCR significantly surpasses SPTS-V2 on TotalText, by 1.6\% without a dictionary and by 0.1\% with a ``full" dictionary.

\textbf{ICDAR2015: Multi-oriented Text}. To evaluate the robustness of our method for multi-oriented text, we conduct experiments on ICDAR2015, with the results shown in Table \ref{tab:results}. Our method outperforms the previous single-point methods across all dictionary settings. Notably, in the strong dictionary setting, InstructOCR achieves an Hmean of 82.5\%, and after adding more training data, it reaches 87.5\%.

\textbf{ICDAR2013: Horizontal text}. To further compare with point-based methods, we conduct experimental comparisons on ICDAR2013, which already has high baseline metrics, resulting in relatively smaller improvements. In the weak dictionary setting, InstructOCR achieves an Hmean of 92.4\%, which is 0.6\% higher than SPTS-v2.

\textbf{Discussion}.
InstructOCR endows the model with the ability to recognize scene text and seamlessly integrate it with the visual context. It is evident that the incorporation of human language instructions indeed significantly aids the model's understanding of text within images. As a result, compared to the  SPTS-V1, there has been a substantial improvement in metrics without a dictionary. Specifically, the improvements are: +2.9\% on the Total-Text dataset, +0.3\% on the ICDAR2013 dataset, and +6.3\% on the ICDAR2015 dataset. This also validates our hypothesis that aligning human language instructions with visual text enhances accuracy.



\begin{table}[h]
  \centering 
  \begin{tabular}{ccccc}
    \toprule
    Methods & Param & Data &TextVQA & ST-VQA  \\
    \midrule
    TAP & 160M & 1.5M& 54.7  & 59.8 \\
    GIT$_{Large}$  & 347M &20M & 37.5 & 44.6 \\
    PreSTU & 278M &13M& 54.5 & 62.6 \\
    WSCOB & 202M &12M &56.2 &62.6 \\
    \hline
    LLaVA1.5 & 7B &1.2M&38.7& 38.1 \\
    InternVL & 8B &4.98B&59.8& 62.2  \\
    Monkey & 8B &1.44M&64.3 & 54.7 \\
    Qwen-VL & 8B & 1.4B & 63.8 & 55.9 \\
    \hline
    InstructOCR & 78M &0.2M &42.0 & 45.8\\
    \bottomrule
  \end{tabular}
    \caption{The public benchmark of TAP~\cite{yang2021tap}, GIT$_{Large}$~\cite{wang2022git}, PreSTU~\cite{kil2023prestu} WSCOB~\cite{kim2023scob}, LLaVA1.5~\cite{liu2023llava},InternVL~\cite{chen2024internvl}, Monkey~\cite{li2024monkey} and Qwen-VL~\cite{Qwen-VL} on TextVQA (acc)~\cite{singh2019towards} and ST-VQA (ANLS)~\cite{biten2019scene} for scene text VQA.}
\label{tab:vqa_benchmark}
\end{table}

\begin{table}[h]
\setlength{\tabcolsep}{4.5pt}
  \begin{tabular}{ccccccc}
    \toprule
    \multicolumn{2}{c}{Text Encoder} & INS & \multicolumn{4}{c}{ICDAR2015} \\
    \cmidrule(lr){1-2} \cmidrule(lr){4-7}
     W & T & & S & W & G & None \\
    \midrule
     \ding{56} & \ding{56} & \ding{56} & 86.5 & 82.7 & 79.1 & 77.1\\
     \ding{51} & \ding{56} & \ding{56} & 86.9 & 82.8 & 79.2  & 77.3\\
     \ding{51} & \ding{51} & \ding{56} & 87.1 & 83.4 & 80.6 & 78.8\\
     \ding{51} & \ding{51} & \ding{51} & \textbf{87.5} & \textbf{84.2} & \textbf{80.6} & \textbf{78.9} \\
      &        &            & (\textbf{+1.0}) & (\textbf{+1.5}) & (\textbf{+1.5}) & (\textbf{+1.8}) \\
    \bottomrule
  \end{tabular}
  \caption{Ablation study of our proposed components on ICDAR2015, ``W", ``T", and ``INS" refer to including a text encoder in end-to-end training, training the text encoder specifically for end-to-end training, and incorporating instructions into the end-to-end training, respectively. ‘S’, ‘W’, ‘G’, and ‘None’ represent recognition with ‘Strong’, ‘Weak’, and ‘Generic’ lexicons and without lexicons, respectively}
\label{tab:instruct_desc}
\end{table}

\subsection{Applicability to Scene-Text VQA}
In this section, we further explore other scene-text related domains (Table \ref{tab:vqa_benchmark}). We show that InstructOCR is also applicable on VQA tasks with a considerable accuracy of $42.0\%$ on TextVQA and $45.8\%$ on ST-VQA. While apples-to-apples comparison is difficult due to different data and parameter sizes, we emphasize the applicability to the VQA task. Specifically, most of the recent works utilize strong backbones such as ViT~\cite{dosovitskiy2020image} and large language models such as T5$_{large}$~\cite{raffel2020exploring}, while ours adopt ResNet-50 and BERT. Our model has, to the best of our knowledge, the least number of parameters ($78M$) among the similar levels of VQA performance, whereas other approaches range from hundreds of millions to even billions of parameters.
It is well known that the VQA performance goes up as more pre-training data is included. The aforementioned methods employ up to millions of text-image pairs, not to mention the multimodal large language models that utilize various forms of large-scale data. Our model is not specifically tailored for VQA tasks and is trained only on 0.2M scene text images. These promising results show the utility and applicability of InstructOCR on image understanding. Our research further extends the boundaries of small-scale models in the VQA task beyond previous limits.

\subsection{Ablation Studies}

We conduct a thorough analysis to understand the individual contributions of each component in our framework, with a particular focus on the effectiveness of the VQA tasks.

\begin{figure*}[t]
  \centering
  \includegraphics[width=0.90\textwidth]{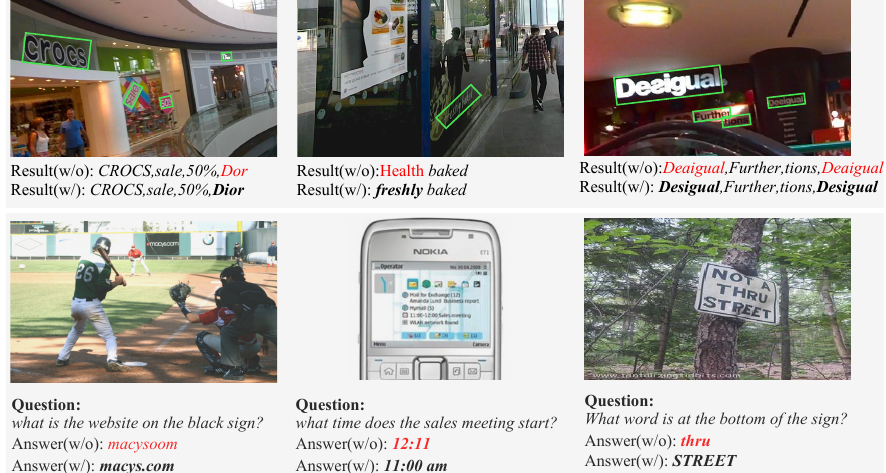}
  \caption{Visualization of correct recognition results on the scene text spotting and VQA datasets after incorporating instructions. The first row corresponds to the scene text spotting dataset, while the second row corresponds to the VQA dataset. ``W/" and ``W/O" indicate whether OCR training was conducted with or without instructions, respectively.}
  \label{figvqa}
\end{figure*}

\subsubsection{Impact of Module Integration}

In this section, we perform a comparative analysis to evaluate the effects of several proposed modules. Prior to validating the effectiveness of instructions, we assess the impact of incorporating a text encoder into the model. We compare the model's performance without a text encoder to its performance with an included and frozen text encoder. We then evaluate the performance metrics when the text encoder is integrated and trained throughout the entire process. The results indicate that the inclusion and active training of the text encoder significantly enhance the capabilities of InstructOCR. Finally, we incorporate instructions, observing further improvements in performance metrics. As shown in Table \ref{tab:instruct_desc}, in the strong dictionary setting, the inclusion of text encoder results in a 1.0\% improvement, while the addition of instructions leads to a further increase of 0.4\%. The experimental results demonstrate that incorporating a text encoder enables the model to more effectively comprehend text within images, and the addition of instructions can further facilitate this. Figure \ref{figvqa} presents some examples where the results became correct after incorporating human language instructions.

\subsubsection{The performance of InstructOCR on the VQA task}
In this section, we conduct experiments to verify the effectiveness of our model in VQA tasks. Since our model is pretrained for scene text spotting, we choose the ST-VQA and TextVQA datasets, which relate to natural scene images. We hypothesize that our framework can be transferred to VQA tasks and that pretraining on text spotting is beneficial for downstream VQA tasks. Table \ref{tab:test_results} supports this argument. Significant improvements are generally observed in VQA tasks when pretraining on text spotting is applied. This is likely because accurate character recognition is a prerequisite for better text understanding, and VQA is closely related to OCR as most answers exist in images containing text. Moreover, the integration of our proposed instructions leads to enhanced image and text understanding, resulting in an improvement of  2.1\% on ST-VQA datasets and 2.6\% on TextVQA datasets. We think that human language instructions during the pre-training stage are beneficial to understanding capability, thereby enabling it to better adapt to more complex understanding tasks in downstream applications. Figure \ref{figvqa} demonstrates the impact of with instructions and without instructions on the output results.

\begin{table}
  \centering 

  \begin{tabular}{cccccc}
    \toprule
    Exp &  W/O & W & INS & ST-VQA & TextVQA \\
    \midrule
     1 & \ding{51} &  &  & 6.7 & 5.6 \\
    2 &  & \ding{51} &  & 43.7 & 39.4 \\
     3 &  & \ding{51} & \ding{51} & 45.8 & 42.0 \\
     &  &  &  & (\textbf{+2.1}) & (\textbf{+2.6}) \\
    \bottomrule
  \end{tabular}
    \caption{Test results for different methods on ST-VQA (ANLS) and TextVQA (Acc.) datasets. “W/O”, “W”, and “INS” refer to training the model without 0.2M OCR data pre-training, training the model with 0.2M OCR data pre-training, training the model with instructions, respectively.}
\label{tab:test_results}
\end{table}

\section{Limitation}

Owing to the VQA task requiring a large amount of data for pre-training and instruction fine-tuning, and different methods using inconsistent data amounts that make it impossible to align data volume. We only use data from the scene text spotting task for pre-training and only use the ST-VQA and TextVQA datasets for instruction fine-tuning. We do not add a large amount of data to demonstrate the best performance of InstructOCR on the VQA task.

\section{Conclusion}

 In this paper, we propose InstructOCR, a novel instruction-based scene text spotting model that leverages instructions to enhance the understanding of text within images. The model integrates a text encoder that processes instructions and a visual-text fusion module that combines image features with these instructions to guide the decoding process. With the introduction of human language instructions and our meticulously designed instruction set based on text attributes, InstructOCR demonstrates an extraordinary ability to comprehend and process textual information within natural scenes, indicating the benefits of aligning human language instructions with visual text for OCR tasks.   Future research will explore increasing training data and incorporating more instructions, thereby enabling the model to achieve higher metrics in VQA tasks and address a broader range of OCR challenges.

\bibliography{aaai25}

\end{document}